\documentclass{article}

\usepackage[nonatbib,final]{neurips_2022_hill}

\usepackage[utf8]{inputenc} %
\usepackage[T1]{fontenc}    %
\usepackage[hidelinks]{hyperref}       %
\usepackage{url}            %
\usepackage{booktabs}       %
\usepackage{amsfonts}       %
\usepackage{nicefrac}       %
\usepackage{microtype}      %
\usepackage{xcolor}         %

\usepackage{graphicx,wrapfig} %
\usepackage{subcaption}
\usepackage{algorithm}
\usepackage{algpseudocode}
\algtext*{EndIf}
\algtext*{EndLoop}
\algtext*{Until}

\title{
Continually Learned Pavlovian Signalling Without Forgetting for Human-in-the-Loop Robotic Control
}

\author{Adam S. R. Parker*, Michael R. Dawson, and Patrick M. Pilarski\\
University of Alberta \& Alberta Machine Intelligence Institute (Amii)\\
\ *{\tt asparker@ualberta.ca}
}

\begin{document}

\maketitle

\begin{abstract}
Artificial limbs are sophisticated devices to assist people with tasks of daily living. Despite advanced robotic prostheses demonstrating similar motion capabilities to biological limbs, users report them difficult and non-intuitive to use. Providing more effective feedback from the device to the user has therefore become a topic of increased interest. In particular, prediction learning methods from the field of reinforcement learning---specifically, an approach termed Pavlovian signalling---have been proposed as one approach for better modulating feedback in prostheses since they can adapt during continuous use. One challenge identified in these learning methods is that they can forget previously learned predictions when a user begins to successfully act upon delivered feedback. The present work directly addresses this challenge, contributing new evidence on the impact of algorithmic choices, such as on- or off-policy methods and representation choices, on the Pavlovian signalling from a machine to a user during their control of a robotic arm. Two conditions of algorithmic differences were studied using different scenarios of controlling a robotic arm: an automated motion system and human participant piloting. Contrary to expectations, off-policy learning did not provide the expected solution to the forgetting problem. We instead identified beneficial properties of a look-ahead state representation that made existing approaches able to learn (and not forget) predictions in support of Pavlovian signalling. This work therefore contributes new insight into the challenges of providing learned predictive feedback from a prosthetic device, and demonstrates avenues for more dynamic signalling in future human-machine interactions.
\end{abstract}

\section{Introduction}
There are many sophisticated devices designed for people who have lost an upper limb to assist them in their daily lives. These devices, prosthetic arms and hands, can take the form of advanced robotic limbs capable of mimicking many, if not all, of the degrees of freedom of a biological limb. Despite the potential of these technologies because of difficulties in control and general use they are sometimes abandoned and therefore unable to fulfill their function \cite{Biddiss2007, Smail2021}. A prosthetic limb is intended for frequent use and close collaboration with the human user; such devices are attached to the body and are intended to be considered a part of the user. The intended closeness of the connection is what makes this particular case of human-machine interaction interesting, and challenging. 

Machine learning has been actively pursued for some time as a way of improving the control of prosthetic upper limbs in various ways, many of which are outlined by Castellini et al. \cite{Castellini2014}. Pattern recognition, for example, learns to associate patterns of muscle activation, often read by surface electromyography (EMG) electrodes, with motions of the prosthesis \cite{Castellini2014, Scheme2011}. Because the connection between the control signal coming from the user and the motions of the device are learned, pattern recognition allows individually tailored solutions for users, which is a highly desirable trait in rehabilitation medicine \cite{Castellini2014}. These methods require training offline however, which means that if there is a change to the user's body or the way they generate control signals the system needs to be retrained to resume proper functionality. A more experimental technique is adaptive switching \cite{Edwards2016}, which learns in real-time to re-order a list of joints that the user has to sequentially switch through to access different movements on a multi-joint prosthesis. This significantly reduces the amount of switching effort required compared to the more common approach of using a fixed list.

\begin{figure}[b!]
    \centering
    \includegraphics[height=0.27\linewidth]{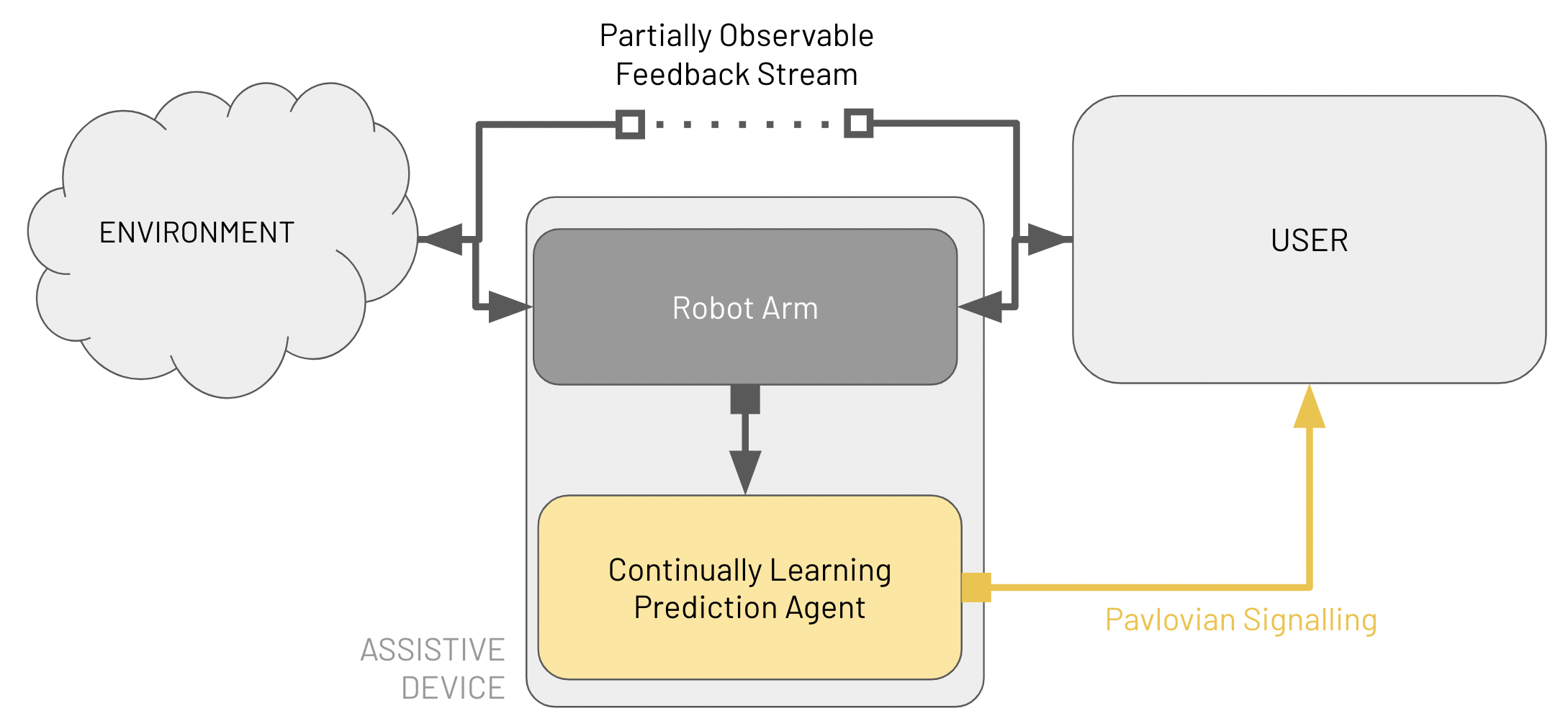}
    \hfil
    \includegraphics[height=0.27\linewidth]{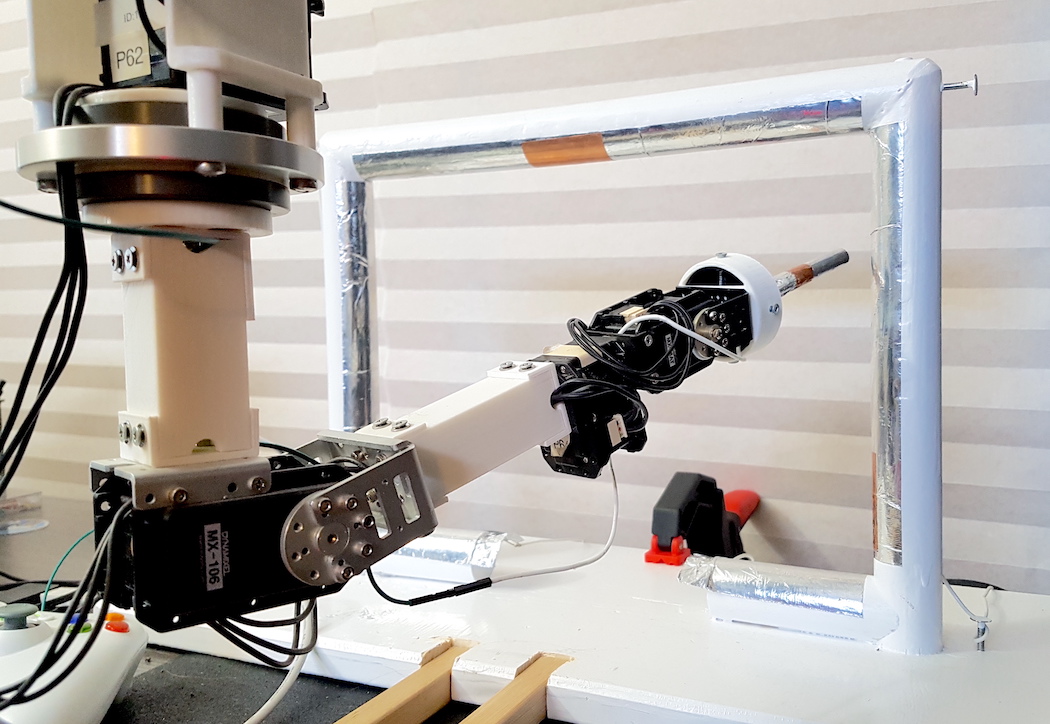}
    \caption{Schematic of the human-machine-environment interaction loop and photo of the physical system: a person interacts with a partially observable environment by way of an assistive machine, wherein their ability to perceive relevant decision-making information is limited in both time and space; as part of their interactions with the environment via the device, a continual prediction learning machine provides signals that allow timely action based on impending changes in the environment, with enough time for the person to process and act on these signals and their current state. 
    }
    \label{fig:scematic}
\end{figure}

Machine learning methods like those used for adaptive switching have the potential to improve upper limb prostheses and rehabilitation in general. Specifically, temporal-difference (TD) learning approaches \cite{Sutton1988} learn in real-time as they are used, and are able to approximate general value functions (GVFs \cite{Sutton2011})---predictions of signals of interest from a robot, human, or environment \cite{Pilarski2013}. GVFs ability to be learned via TD in real-time gives them the potential to adapt and grow as the user does \cite{Pilarski2013}. This could allow users to develop their skills and abilities as they more naturally would, rather than having to learn and compensate around preset capabilities of the device they are using. The ability of reinforcement learning techniques to adapt also unlocks the potential for personalized user solutions to be found by the system, rather than having designers customize every device for every user \cite{Castellini2014}. While machine learning has been studied extensively for control improvement in prostheses, {\em it's applications to providing feedback from the prosthesis to the user remain under explored.}

Feedback from a limb to the user can be an important aspect of control as well as user experience. Sensory feedback has been shown in some studies to improve user outcomes, as it closes the loop between the user's control and the device's execution. The comprehensive review by Schofield et al. \cite{Schofield2014} covers many methods and modalities of providing feedback. Non-biological feedback, such as sound relating to direction of motion, can also help with the development of internal models, which are hypothesized as being what the central nervous system uses feedback to develop and are potentially responsible for the performance gains when feedback is available \cite{Shehata2018}. Feedback of this type can be thought of less as feedback in the sense one might have from a biological limb and more along the lines of signalling. To date most of the explorations into providing feedback have used a fixed mapping between the sensor and the feedback device that does not change or adapt over time.

There are other domains where interactions between a human and a device have shown adaptive feedback to be helpful to the human \cite{Crandall2018}. These domains are beginning to include machine learning, and explore how it might signal a human user to assist in accomplishing a task the human would otherwise struggle with \cite{Crandall2018,Brenneis2021}. Early explorations of this kind used what has been recently termed as \emph{Pavlovian signalling}: a fixed signalling response (a token) emitted by one agent---or part of an agent---to another in response to the magnitude of a learned prediction about the environment or a task \cite{Pilarski2022} (Fig. \ref{fig:scematic}). Studies have been done where a machine agent is learning about a task and providing the human user with a signal they then use to successfully complete the task \cite{Brenneis2021,Pilarski2022,Parker2019}. In this way the human and the machine can be seen as partners in accomplishing the task, and effective, reliable, communication is key.

An earlier study by Parker et al.\ took a very simple approach to Pavlovian signalling in human machine interaction, specifically the task of navigating a robot arm in a confined workspace without touching the walls \cite{Parker2019}. The paper used TD learning---an on-policy method, i.e., one that learns about the actual policy or behaviour being followed---to acquire GVF predictions, but found that learning had to be done during a training session and frozen during a trial. This was because when the machine agent was succeeding at helping the human avoid contacts with the work space, the machine agent was also avoiding those contacts and so unlearning them or ``forgetting" they were there. While the results showed that the machine agent could learn something that assisted the human users in accomplishing the task, being unable to learn continually in real-time undermines the benefits of using continual learning techniques like TD learning. It was theorized that off-policy learning would overcome this limitation, i.e., that forgetting during continual learning could be mitigated by the use of learning algorithms that learn about a target policy different from the policy being followed, capturing ``what if'' style predictions \cite{Sutton2011} about user behaviour even when the user is performing well in a task or setting. 
In this study, we extend prior work to contribute a detailed examination of hypothesized differences between on- and off-policy TD learning in the context of forgetting during Pavlovian signalling in ongoing robotic control. As this work is interested in exploring the application of machine reinforcement learning techniques in a rehabilitation setting in order to investigate solutions that adapt with the user over time to become individual, personalized, user experiences, the findings on the single human participant are not meant to be generalized. They are, however, worth bearing in mind as if something doesn't work for even one person, then this impacts the ability of the technique to become a personal solution for at least that individual \cite{Mook1983}.

\section{Methods}

\subsection{Experimental Setup}
The robot arm used for these experiments was the Bento Arm \cite{Dawson2014}. This arm is a 5-degree-of-freedom open-source 3D printed device that uses servo motors for actuation at the shoulder, elbow, wrist, and hand as shown in Fig. \ \ref{fig:scematic}. For this work, the Bento Arm hand was replaced with a rod wrapped in conductive tape. The work space for the arm was a square interaction region constructed of wood and lined with conductive tape, which is also shown in Fig. \ref{fig:scematic}, 
with these conductive elements all connected to analog inputs of an Arduino Leonardo.
For human-participant interaction, the thumb-stick of an Xbox 360 controller was used to control the shoulder joint of the arm; built-in vibration of the controller was used to signal the participant that a trial was complete. Contact, or the prediction of contact, between the conductive rod and the workspace was signalled to the participant using a sound effect delivered over a set of noise-cancelling headphones with ambient white noise playing in them. The Bento Arm, Xbox 360 controller, and Arduino were connected to a laptop via USB. The laptop ran a modified version of the control software brachI/Oplexus \cite{Dawson2020}. This software package designed for use with the Bento Arm with built-in functionality for multiple control sources, as well as the ability to map control inputs to the Bento Arm.

\subsection{Procedure}

Two different settings were used to study the impacts of the algorithmic decisions made in this work: machine-machine control interactions and single case of human-in-the-loop control. For machine-machine interactions\, which provided a clear baseline for algorithmic comparison, we used a version of brachI/Oplexus that included a motion sequencer feature that allowed users to program step-by-step arm motions, where upon reaching a set position the arm automatically transitions to the next movement in a list. This was modified for the autonomous trials to use the signal from the Arduino's analog inputs to move on to the next motion on the list. After a contact or prediction were observed to be above a threshold, the next motion was to back-off a pre-determined amount, and then move in the opposite direction. One complete motion involved moving to the right, backing off the pre-determined amount, moving to the left, then backing off again. For human-in-the-loop comparisons, a single human participant was recruited, who provided informed consent and the study was approved by the Research Ethics Board of the University of Alberta (Pro00085727).

The goal in both settings was to control the robot arm to move from side-to-side without making contact with the workspace edge, following the protocol established in Parker et al. \cite{Parker2019}. Both the human participant and automotion sequence moved the robot's shoulder actuator from left to right and back again within the workspace. Signalling tokens relating to contact with the workspace and predicted contact (described below) were provided as feedback in both the human and machine case (Fig. \ref{fig:scematic}). For the human participant, the task was made partially observable for them by having them face away from the workspace while being sound isolated in noise-cancelling headphones with background white noise---they needed to rely completely on the feedback system to perceive the outcomes of their robot control commands. Similarly, the automotion controller was given no information about the workspace other than the tokens provided through feedback signalling. One trial consisted of 50 back-and-forth motions. Each learning technique was run for five trials by automotion and the human participant did three trials of each of the algorithms they tested.

\subsection{Prediction Learning Algorithms}

Algorithm~\ref{alg:flow} presents the main loop of the Pavlovian signalling process that occurred during user-robot interaction: observations from the robot and workspace were sampled and used to generate predictions; should these predictions rise above a threshold (or actual contact be made), a token was generated and sent to the user of the robot (automotion or human) that informed and/or changed their control of the system (c.f., \cite{Pilarski2022}). For the human participant, this took the form of an audio cue proactively alerting them to contact with the workspace; for the automotion case, this took the form of the robot arm backing slightly away and changing direction. all predictions were then updated according to the information obtained during the time step. The threshold empirically chosen (400) and constant across all conditions.

Predictions themselves were learned via temporal-difference (TD) reinforcement learning techniques \cite{Sutton1988,Sutton2011,Sutton2018} to generate predictions about the expectation of future contact \cite{Parker2019}. These learning algorithms learned from experienced data in real time during each trial; there was no training period before any of the trials. Three different implementations of GVF learning approaches were examined and compared for their ability to avoid contact events, and continue to avoid contact events over repeated motions: TD($\lambda$), GTD($\lambda$)\cite{Sutton2018}, and a variant we term look-ahead TD($\lambda$). All three approaches used the same representation which consisted of the shoulder position, shoulder velocity, elbow position, and elbow velocity as reported by the servos responsible for motion of the Bento Arm. To construct the representation the allowed range of motion and velocities of the shoulder and elbow were normalized to produce a value between zero and one with zero being the left-most limit of the shoulder and bottom-most limit of the elbow, and one being the right-most limit of the shoulder and upper-most limit of the elbow. This was then divided into 32 evenly spaced sections, or bins, on two axes, one for position and one for velocity---an  implementation of the function approximation method know as tile coding \cite{Sutton2018}.

\begin{wrapfigure}[18]{r}{0.45\textwidth}
\begin{minipage}{0.45\textwidth}
\vspace{-2.1em}
\begin{algorithm}[H]
\caption{Pavlovian Signalling}\label{alg:flow}
\begin{algorithmic}
\State \textbf{set} {\em thresh} for Pavlovian signalling
\Loop
\State \textbf{observe} {\em contact}
\State \textbf{observe} GVF {\em prediction} value

\If{{\em contact} or {\em prediction} $>$ {\em thresh}}
\State generate {\em signalling token}
\EndIf
\If{{\em signalling token}}
\If{human trial}
\State signal with audible feedback
\EndIf
\If{automated motion}
\State back-off motion
\State change direction
\EndIf
\State clear {\em signalling token}
\EndIf
\State update all {\em predictions}
\EndLoop
\end{algorithmic}
\end{algorithm}
\end{minipage}
\end{wrapfigure}

Following on work by Parker et al. \cite{Parker2019}, the first GVF-learning used was the TD($\lambda$) algorithm, which was tested using two different $\lambda$ values, 0 and 0.9. In this algorithm, a temporal-difference error $\delta$ is computed based on an observed cumulant $C$, and the difference between the learned expected value for the present state (the inner product of a weight vector $\vec{w}$ and the function approximated state $\vec{x}(S)$ based on robot arm measurements $S$ as noted above, discounted by $\gamma$) and the value of the last state $\vec{w}^\mathsf{T}\vec{x}(S_{Last})$. This error is used alongside a step size $\alpha$ and replacing eligibility traces $\vec{e}$ with decay rate $\lambda\gamma$ to update $\vec{w}$.  Parker et al. \cite{Parker2019} used a TD algorithm with $\lambda=0$, so this was re-examined here to compare to using a $\lambda$ value above zero.

As our second approach, we introduced a comparison with gradient temporal-difference learning (GTD) methods \cite{Sutton2018}. GTD($\lambda$) is an off-policy algorithm; an off-policy algorithm learns about a target policy $\pi_t$ from samples generated by a possibly unrelated behavior policy. The observed overlap between target and behaviour policies is specified by a parameter $\rho$, and is able to learn stably in this setting thanks to a separate set of weights $\vec{u}$ and learning rate $\beta$. 
For the off-policy comparison, the system learned two GVFs: one for each direction of motion of the shoulder actuator. For this study, $\rho$ was always either zero or one; $\rho$ was one for the GVF responsible for learning the direction the servo was currently moving in, and 0 for the other direction. 

The final approach we explored was TD($\lambda$) but utilizing a fixed look-ahead in the computation of the observation $S$, and is therefore termed ``look-ahead TD". Rather than learning and acting on the same time step, this method learned on the active time step but used the prediction for a fixed number of tilecoding bins ahead of where the arm actively was learning to use to signal either the human participant or the automated motion controller. %
Algorithms for both TD($\lambda$) and GTD($\lambda$) are according to the standard implementation and can be found in the supplementary materials.

\section{Results}

\begin{wrapfigure}[44]{r}{0.5\textwidth}
\vspace{-5em}
    \centering
    \includegraphics[width=0.95\linewidth]{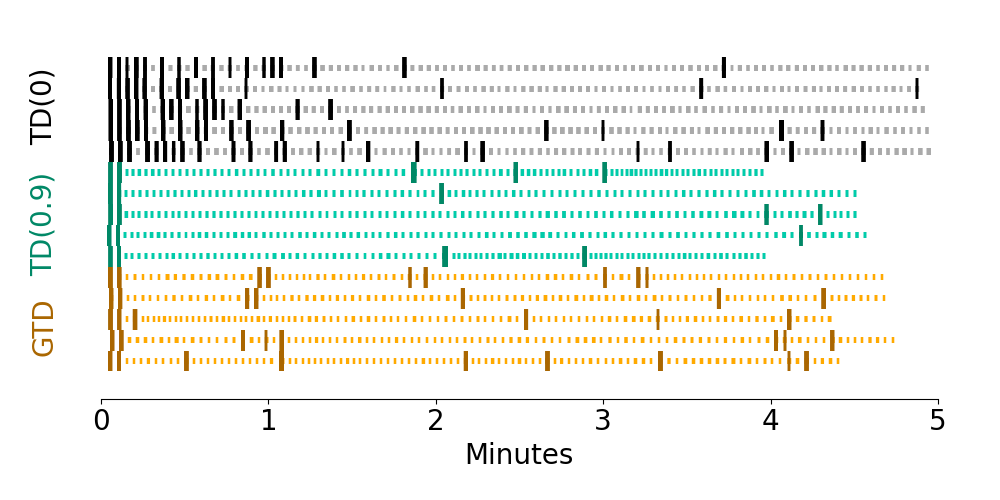}    
    \caption{Contacts in the automatic motion case over time for all 5 trials for different algorithm settings: (black top) TD(0) (teal, middle) TD(0.9) (orange, bottom) GTD. Large ticks indicate contact, small ticks indicate generation of signalling tokens based on learned predictions.}
    \label{fig:results1}

    \includegraphics[width=0.95\linewidth]{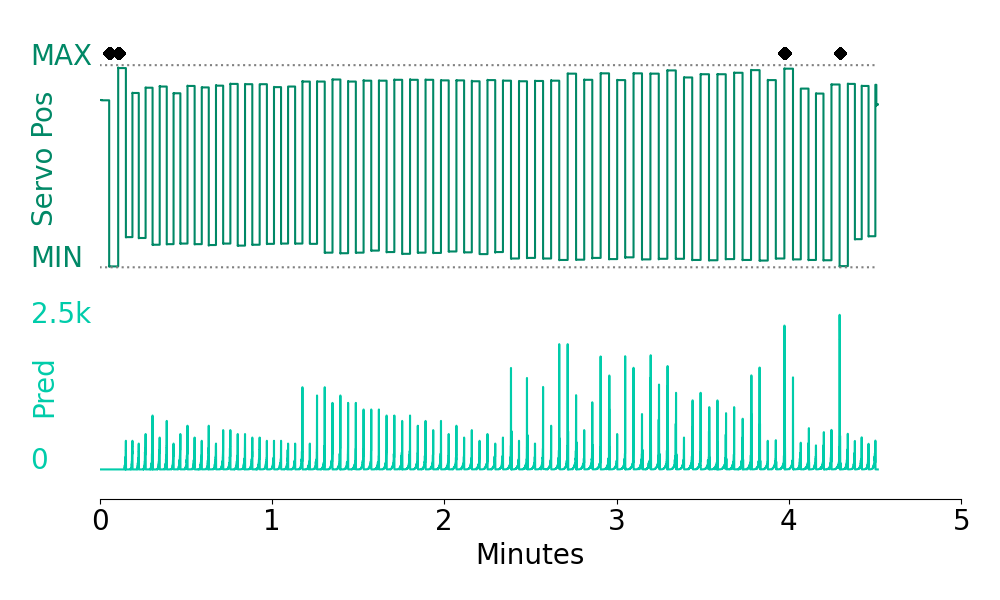}
    (a) \includegraphics[width=0.95\linewidth]{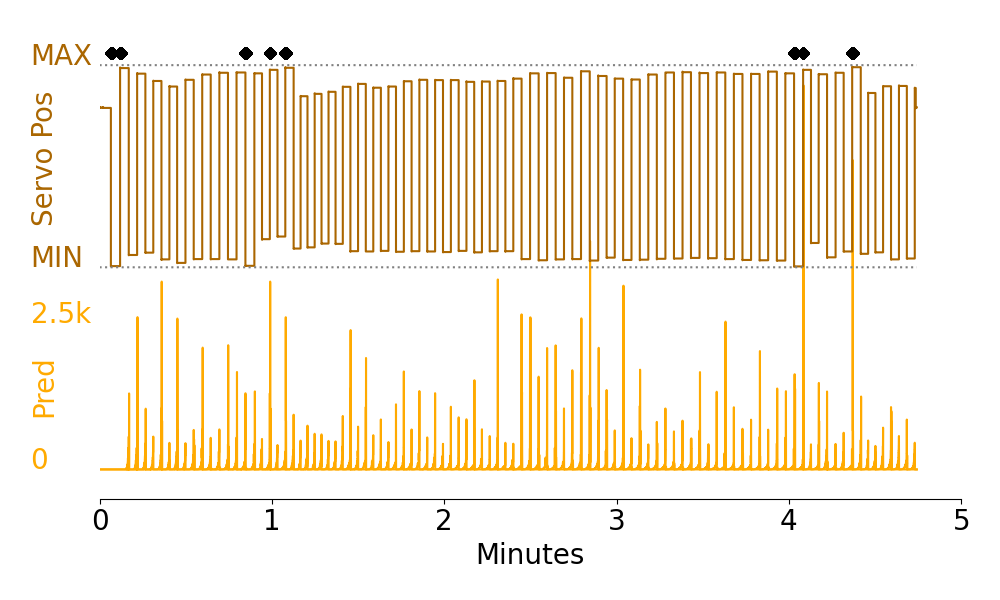} (b)\\
    
    \caption{Contacts (black diamonds), motion (top) and prediction (bottom) for exemplar trials of (a) TD-lambda and (b) GTD to show the motion getting closer to the contact positions, marked by the dashed lines around the position, and the predictions diminishing, and spiking when a new position extreme is moved into. %
    }
    \label{fig:results2}
\end{wrapfigure}

Figure \ref{fig:results1} shows the contact events, the dark lines, and the prediction going above the threshold value, the light lines. The three initial algorithms for the study, TD(0), TD(0.9) and GTD. TD(0) requires the most contacts to learn the boundaries of the workspace in the first place, and continues to make contact frequently. GTD initially learns fast, but continues to make frequent, though less frequent, contact. TD(0.9) also learns to avoid contact quickly, but makes contact much less frequently over the duration of each trial. %
In Fig. \ref{fig:results2}, we can see the contacts as they relate to the position of the shoulder for TD(0.9) and GTD. The position, shown in the top part of each plot is the location where the servos reported stopping, which typically occurred during changes of direction. 
Primarily of note is how the extremes of position can be seen expanding by bin sizes of position until contact, the location of which is shown by the dashed envelope around the position. After a contact is made the motion extremes then retract and the prediction, shown on the bottom part of each plot spikes. This spike is also visible when the position extremes expand, and is likely the result of visiting a new bin that has already learned about contact and not been visited in recent motions.
This is more shown in more detail the supplementary material Fig.\ S1, where the prediction can be seen diminishing over repeated motions until it drops below the threshold and contact occurs. %

TD(0.9) compared with look-ahead TD is shown in Fig. \ref{fig:results3}. Figure \ref{fig:results3}a specifically compares contacts over the duration of trials between TD(0.9) alone and TD(0.9) with a look-ahead TD of one and two bins. The one bin look-ahead TD did occasionally make contact, which should not be possible since the predictions on the timesteps being used to signal are not being updated on the same timestep they are used. It can be seen in Fig. \ref{fig:results3}b that the one bin look-ahead TD prediction does not diminish, but a contact happens anyway. This causes the prediction to increase and motion to restrict further. This happens again on the other extreme of motion. Figure \ref{fig:results3}c shows the prediction and motion for two bin look-ahead TD. We can still see the shoulder position at reported motion stops not being perfectly steady as we would expect from robotic motion, but drifting slightly in both extremes. The additional bin prevents these drifts from making contact.

Human participant data is show in Fig. \ref{fig:results4}. The contacts over the trials for TD(0.9) shown in Fig. \ref{fig:results4}a show that TD(0.9) made repeated contacts after over the course of the trials. TD(0.9) with a two bin look ahead also struggled. This is likely a result of additional delays in the human-machine system from the sound used to signal the human having a delay in its onset and the reaction time of the participant. Figure \ref{fig:results4}b shows the predictions for TD(0.9) with the look ahead increased to four bins remains high over the course of the trial, but a single contact was still made in the 50 motions of the trial despite this. The motion that this happened on is one straight shot; it doesn't have any of the pauses before the extreme that some of the motions exhibit. The two bin look-ahead TD shown in Fig. \ref{fig:results4}c shows more passing through the bin where the signal is initially triggered by the prediction and into bins beyond---despite the prediction at the extremes of motion being high, contacts are made.

\begin{figure}[t]
\centering
    \begin{minipage}[t]{0.48\textwidth}
    \centering
    \includegraphics[width=0.95\linewidth]{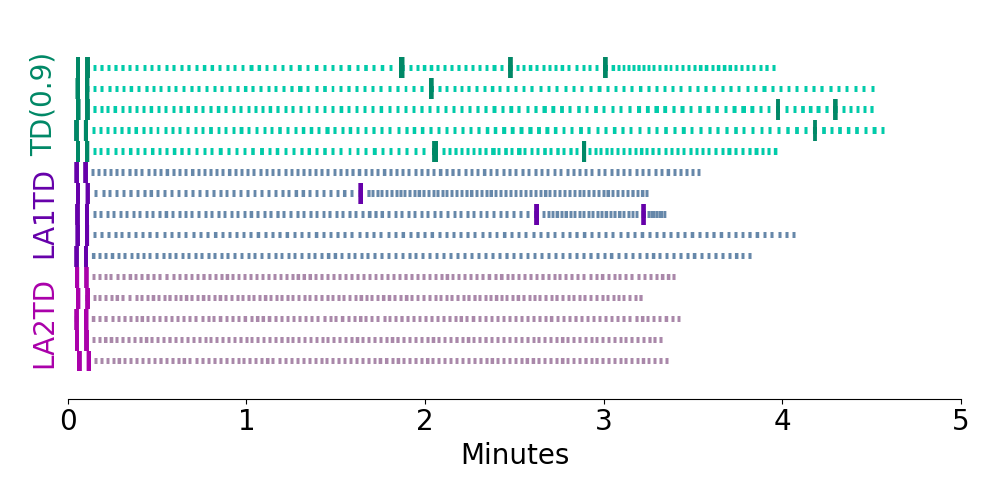}\\
    (a)\\
    \includegraphics[width=0.95\linewidth]{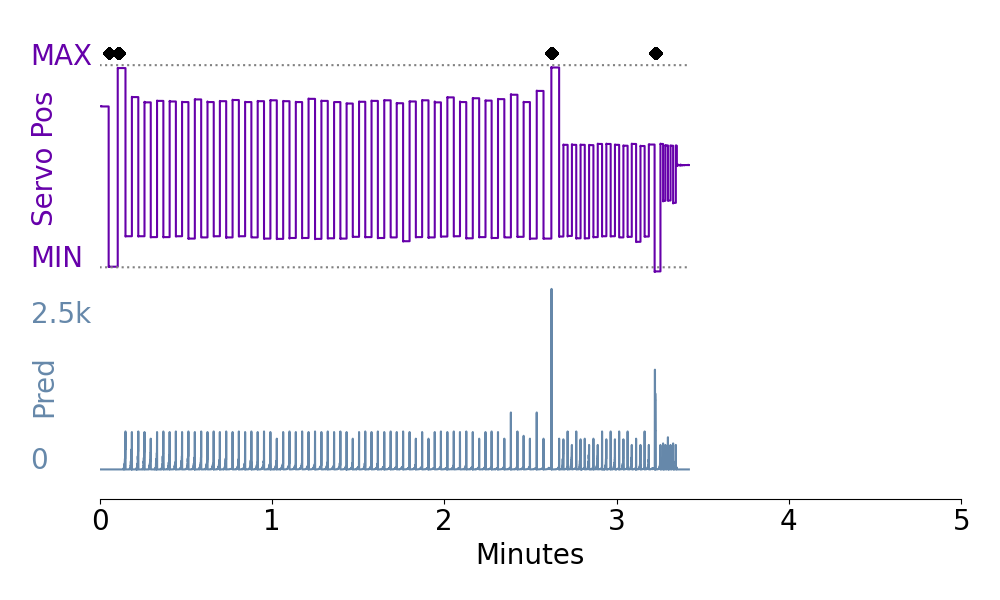}\\
    (b) \\
    \includegraphics[width=0.95\linewidth]{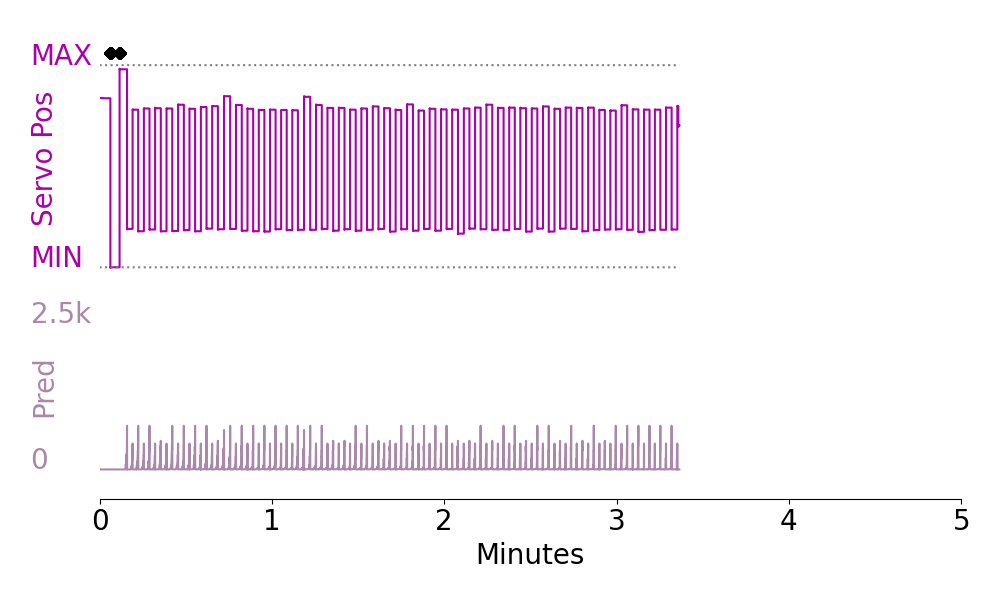}\\
    (c)
    \captionof{figure}{(a) Contacts as in Fig. \ref{fig:results1}, but for TD(0.9), 1 bin look-ahead (LA1TD), and 2 bin look-ahead (LA2TD).
    Contacts, position, and predictions for (b) LA1TD and (c) LA2TD.%
    }
    \label{fig:results3}
    \end{minipage}%
        \hfill
    \begin{minipage}[t]{0.48\textwidth}
    \centering
    \includegraphics[width=0.95\linewidth]{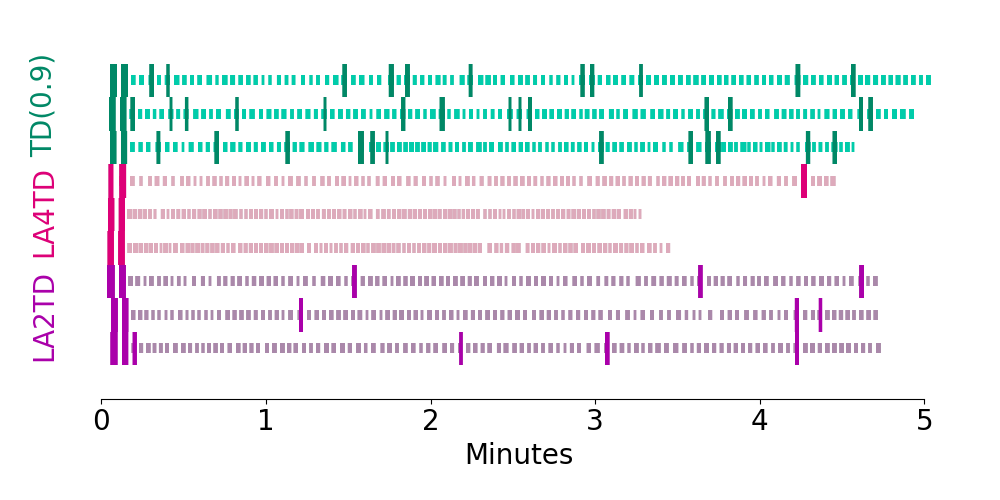}\\
    (a) \\
    \includegraphics[width=0.95\linewidth]{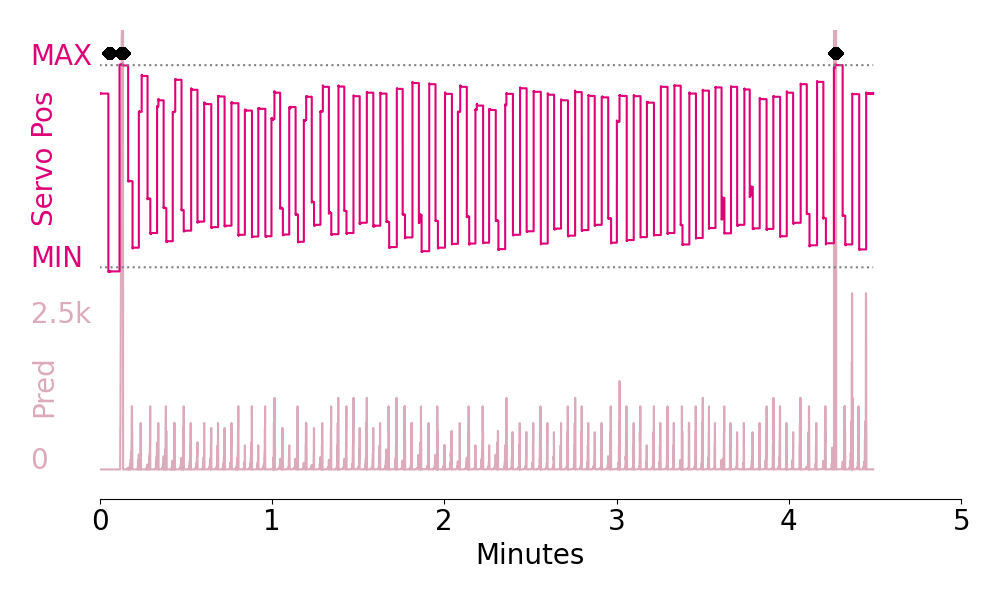}\\
    (b) \\
    \includegraphics[width=0.95\linewidth]{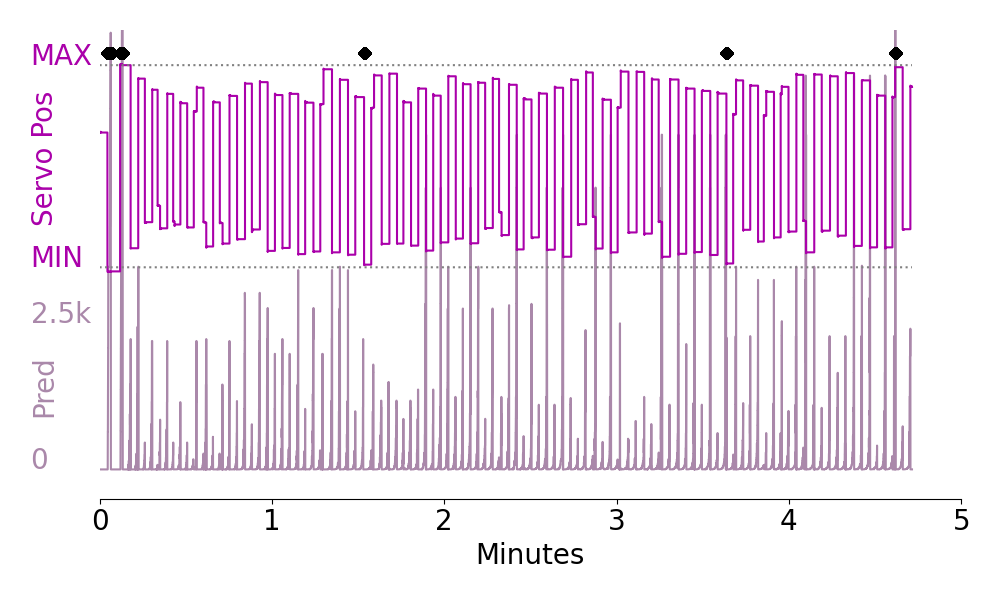}\\
    (c)
    \captionof{figure}{Participant data for (a) contacts as Fig. \ref{fig:results1} all three cases tested with the human participant, and contacts, motion, and prediction(b) LA4TD and (c) LA2TD. %
    }
    \label{fig:results4}
    \end{minipage}
\end{figure}

\section{Discussion}

\textbf{TD Compared to GTD:} In a previous study it was found that a machine learning agent using reinforcement learning techniques can learn something that is of value to the more sophisticated sensorimotor decision making of a human user \cite{Parker2019}. In that study, however, one of the primary advantages of reinforcement learning techniques was not utilized in the trials---the system was not learning while the task was being done. This was because there is a particular challenge involved when success on the task involves no longer encountering the event in the signal space that is trying to be avoided. It was proposed in Parker et al. \cite{Parker2019} that off-policy learning would overcome this. 
That proposal was the first algorithm examined in our present study. Surprisingly, Fig. \ref{fig:results1} clearly shows that the off-policy learning algorithm, GTD in particular, did not overcome the forgetting seen in Parker et al. The closest algorithm to that of the original study used here was TD(0), and Fig. \ref{fig:results1} does show how this method struggles. However it is also clear that GTD did \emph{worse} than TD(0.9). This was with robotic auto-motion, which is relatively consistent when compared to a human interacting with the device. This indicates that a human user would not find the expected success using an off-policy algorithm while the agent is learning online in real time as is the goal.

In order to determine what is happening further exploration of the data was required. 
Figure \ref{fig:results2} suggests some of the challenges faced by GTD.
It seems that part of what was supposed to be the benefit of the off-policy algorithm becomes a challenge; when the system motion stops or changes direction the observations do not impact learning about motion in the direction it was previously travelling. In this case, the agent gets less experience on each motion and learns more slowly. The pattern of diminishing predictions and increasing extremes of position until contact is visible here as well but occur more frequently. This strongly suggests that the benefit we were expecting to see from off-policy learning, not forgetting about where contacts were made when it no longer encountered them, is being at least in part overridden by the lower number of samples seen by each GVF as they are learned.
more detail of the contact events can be seen in supplementary material Fig.\ S1. In particular closer examination of the predictions shows the prediction gradually diminishing over repeated motions in both directions until it is below the threshold and contact is made.

It is important that the machine learning agent tasked with assisting the human does not ``forget" about a feature when success involves avoiding that feature. This places the machine agent in a position where it is at best not providing anything useful to the human other than perhaps some initial priming of the human's expectations of a task, after which the human must ignore the signals from the agent. At worst the agent is damaging the interaction between the human and the device by being unreliable and untrustworthy \cite{Schofield2021,Brenneis2021}. At the same time, the machine agent should be able to forget when it is appropriate to do so, or it risks becoming specialized to a very specific set of circumstances and is not fulfilling a primary advantage of reinforcement learning methods in continual human-machine interaction of real-time adaptability. If the human chooses to ignore a signal from the machine agent, and the workspace shape had changed so that contact was further out, the system should be able to adapt to this change in how the human is doing the task. For these reasons it is worthwhile to pursue reinforcement learning methods to improve human-machine interaction.

\textbf{TD Compared to Look-Ahead TD:} There are several challenges to this task that necessitated the use of a simple domain to attempt to isolate the effects algorithmic differences on learning to avoid some part of a workspace or task, and doing so to reliably generate predictions to aid this avoidance. The avoidance itself poses a challenge for the TD learners. Since these algorithms learn from experience, when they no longer experience something they not only stop learning about it but the experience that is avoided is replaced with more recent experiences. GTD, an off-policy algorithm, was thought to be the solution to this problem, however it encounters challenges as well. It is very possible that some of these challenges are based in the nature of physical systems; the play in the servos, inertia from the motion, and time synchronicity challenges that arise in part as a result of running each part of the system as fast as it is able in order to not miss events of interest. It should be noted that the servos used in the Bento Arm are durable, high precision servos, so while there may be hardware that would assist in overcoming the physical challenges the expense becomes significant. 
Therefore, instead of physical modifications to the arm, we changed how the predictions were being accessed. Until now, the timestep that was being learned on was also being used to generate the prediction token to instruct the system to change directions in order to avoid contact. This was the purpose of the look-ahead TD method; by using the prediction a specified number of bins ahead of the active bin in the state space, 
this method should make it impossible for the predictions being used to signal to diminish since they are not physically visited by the system. 

TD(0.9) was used as the learning algorithm with look-ahead TD, as it learned quickly. 
Of particular note in Fig. \ref{fig:results3}a, the two bin look-ahead made no contacts after an initial one per side. The one bin look-ahead did, on occasion make contact after initial contacts. 
It is clear in Fig. \ref{fig:results3}c that the motion is stable; the extremes of position don't change substantially for the most part, except in a few places where motion stopped in a further bin, but no contact was made. The predictions remain undiminished, at least in their peaks. Interestingly, upon close inspection what happens to the predictions is they decay in the bins being visited, but not the ones being used. This gives them the appearance of a spike, rather than a gradual increase.
The appearance of the motion and prediction in Fig. \ref{fig:results3}b and Fig. \ref{fig:results3}c is very similar until the contacts, at which point Fig. \ref{fig:results3}b subsequent motions in that direction stay further from the previous extreme as the prediction value of all relevant bins, as configured in the GVFs, increases. It is interesting that look-ahead of any number made contact at all. This strengthens the implication that there are more issues facing the system than unlearning the events it stops observing. It is likely the contacts made by the one bin look-ahead can be attributed to previously mentioned physical challenges of the system. That there seems to be no regularity to the contacts caused by this is an extra challenge, that seems as though it can be avoided by increasing the number of bins of look-ahead.

\textbf{Human Participant Outcome:} In order to begin to explore the impact that these algorithmic choices have on the ability of the system to adapt and become an individualized solution in human-machine interaction, a case study was done with a single participant using TD(0.9) and look ahead. The originally pursued off-policy GTD was not included in the human controlled setting as it's problems would only be exacerbated by human use. Figure \ref{fig:results4}a shows that TD(0.9) does not fare as well at reducing the number of contacts with the human participant as it did with automated motion. The two bin look-ahead had fewer repeated contacts, but with the ideal goal being zero contacts after the initial motions it is not performing adequately. Even increasing the look-ahead to four bins resulted in a single contact. This contact happened on a trial that took longer to complete than the other two, which implies that it was covering more distance between extremes and thus was closer to contact the entire trial. 
The motion of the human shown in Figs. \ref{fig:results4}b and \ref{fig:results4}c is, predictably, less regular than that of automated motion. This, coupled with delays in reaction caused by the participants reaction time and a software delay in the onset of the sound signalling the participant of an event, increased the difficulty of the task for the learning agent. Even with four bins of look-ahead there was a contact late into one of the trials.
The look-ahead technique, if nothing else, adds an extra parameter that can be adjusted to assist the success of the learning agent. While look-ahead TD should completely prevent forgetting, even using on-policy algorithms such as TD($\lambda$), the physical factors of the system make it not so simple. If the number of look-ahead bins is not correctly set, over extended periods of time the slippage into adjacent, closer, bins would still lead to forgetting. It is possible, in fact likely, that different representations and even reward functions could be used to accomplish this specific task. However, the choice of representation and cumulant here were made because of they do not use designer knowledge of the task, but rather are grounded in the construction of the robot, specifically the information available from the servos, and signals from the environment alone. 
By grounding the representation and cumulants in the construction of the robot in this way, the system must learn about the environment and task by way of it's own motions and signals. This is important because, if successful, it allows the system to learn about the world in terms that will always be available to it even with different users, tasks, and environments. The use of a single participant for the human portion of these trials is a significant limitation to statements about the generality of this work. However, in this domain, the driving rational of using RL techniques is that they can adapt to an individual user, therefore any single participant where this is found to be not true (as in the present experiment) is worth bearing in mind for future work. For the interested reader, a more advanced proposal for learning predictive look-ahead can be found in the supplementary material (Fig.\ S2).

\vspace{-1em}

\section{Conclusions}

\vspace{-1em}

In this work we investigated a clear prosthetics-motivated example of one of many settings where agents interact with other agents in uncertain via signals that they adapt in real time and through ongoing experience. In addition to its specific contributions to improving feedback from prosthetic limbs by demonstrating the use of Pavlovian signalling in a human-robot arm interaction, this paper provides concrete evidence on how algorithmic differences impact continual temporal-difference learning of approximate general value functions used in feedback and signalling; this work contributed new insight into the importance of on- and off-policy learning choices, predictive representations of state, and function approximation, and how these factors act differently on real-world platforms with and without a human in the loop. As this work is intended to further methods for reinforcement learning techniques to adapt to individual users and provide specific solutions, the findings on the inability of the selected off-policy method to reliably provide signals the human could use is worth keeping in mind moving forward. We therefore expect these findings to support the development of next-generation neuroprostheses and other assistive technology, and more broadly a range of diverse applications where multiagent interaction occurs in complex domains in concert with or as mediated by prediction learning machines that continually learn during ongoing human-in-the-loop interaction.

\section*{Acknowledgments and Disclosure of Funding}
We thank Victoria Collins for making our studies possible. Research was supported by the Natural Sciences and Engineering Research Council of Canada (NSERC), Alberta Innovates, the Sensory Motor Adaptive Rehabilitation Technology (SMART) Network, the Alberta Machine Intelligence Institute (Amii), and the Canada CIFAR AI Chairs program.

\bibliographystyle{ieeetr} %
\bibliography{refs} %

\newpage
\section{Supplementary Material}
\section*{S1\ \ \ Temporal-Difference Learning Algorithms}

\renewcommand{\algorithmicrepeat}{\textbf{on update call}}

\begin{algorithm}
\renewcommand{\thealgorithm}{S\arabic{algorithm}}
\setcounter{algorithm}{0} 
\caption{TD($\lambda$) Update}\label{alg:TD}
\begin{algorithmic}
\State \textbf{set} $\alpha \gets 0.1, \gamma \gets 0.9, \lambda \gets 0$ or $0.9$
\State \textbf{init} $\vec{w}, S, S_{Last}, \vec{e} \gets 0$
\Repeat
    \State \textbf{observe} $C, S$
    \State $\delta \gets C+\gamma\vec{w}^\mathsf{T}\vec{x}(S)-\vec{w}^\mathsf{T}\vec{x}(S_{Last})$
    \State $\vec{e} \gets \mathsf{min}(1, \vec{x}(S_{Last})+\gamma\lambda\vec{e})$
    \State $\vec{w} \gets \vec{w}+\alpha\delta \vec{e}$
    \State $S_{Last} \gets S$
\Until{}
\end{algorithmic}
\end{algorithm}

\begin{algorithm}
\renewcommand{\thealgorithm}{S\arabic{algorithm}}
\caption{GTD($\lambda$) Update}\label{alg:GTD}
\begin{algorithmic}
\State \textbf{set} $\alpha \gets 0.2, \gamma \gets 0.9, \lambda \gets 0.9, \beta \gets 0.01$
\State \textbf{init} $\vec{w}, \vec{u}, S, S_{Last}, \vec{e} \gets 0$
\Repeat
    \State \textbf{observe} $C, S$
    \State $\rho \gets 0$
    \If{behaviour aligns with $\pi_t$}
        \State $\rho \gets 1$
        \EndIf
    \State $\delta \gets C+\gamma\vec{w}^\mathsf{T}\vec{x}(S)-\vec{w}^\mathsf{T}\vec{x}(S_{Last})$
    \State $\vec{e} \gets \rho \cdot \mathsf{min}(1, \vec{x}(S_{Last})+\gamma\lambda\vec{e})$
    \State $\vec{w} \gets \vec{w}+\alpha[\delta \vec{e} - \gamma(1-\lambda)(\vec{e}^\mathsf{T}\vec{u})\vec{x}(S)]$
    \State $\vec{u} \gets \vec{u}+\beta[\delta \vec{e}-(\vec{x}(S_{Last})^\mathsf{T}\vec{u})\vec{x}(S_{Last})]$
    \State $S_{Last} \gets S$
\Until
\end{algorithmic}
\end{algorithm}

\newpage

\section*{S2\ \ \ Extended Results}

\begin{figure}[h]
    \centering
    \renewcommand{\thefigure}{S\arabic{figure}}
    \setcounter{figure}{0} 
    \includegraphics[width=0.6\linewidth]{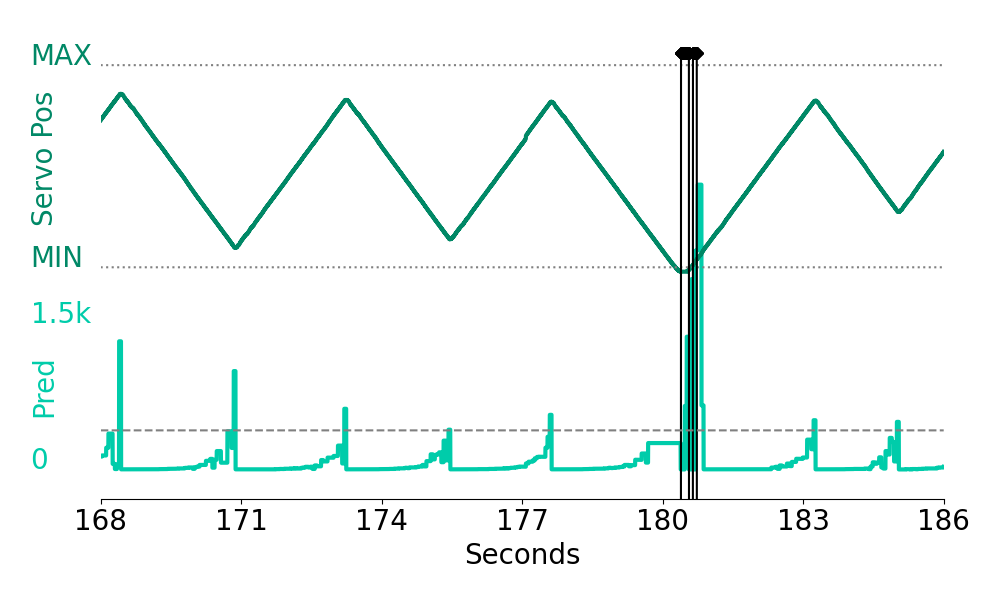}\\
    \caption{Contacts (black), motion (top), and prediction (bottom) for TD-lambda zoomed in to show the point where contact occurs. Dashed line across the prediction is the threshold for token generation. Here the position is as reported by the servo on each timestep.}
    \label{fig:results2b}
\end{figure}

\section*{S3\ \ \ Future Directions: Learning the Predictive Look-ahead}

\begin{figure}[h]
    \centering
    \renewcommand{\thefigure}{S\arabic{figure}}
    \includegraphics[width=0.78\linewidth]{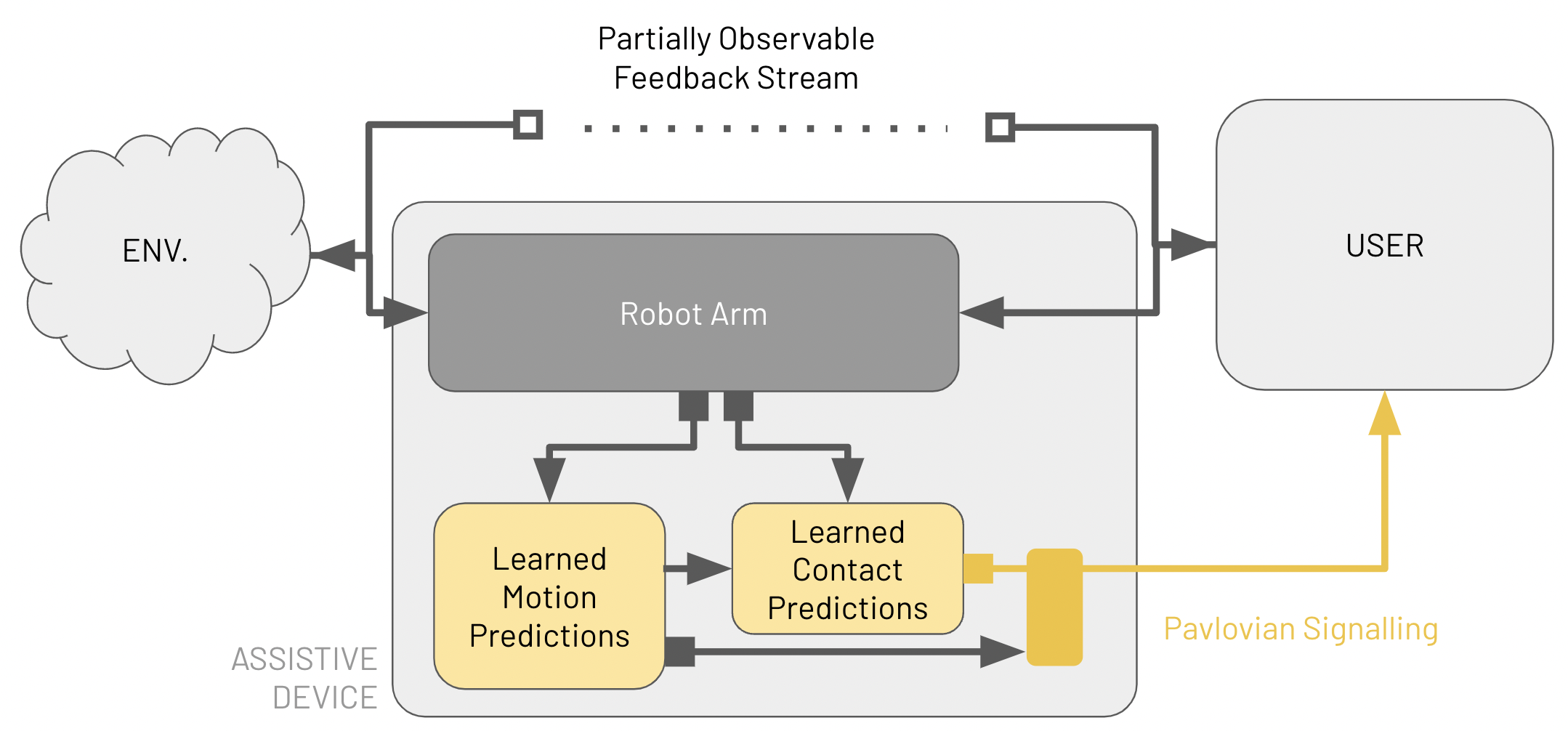}
    \caption{Schematic of more advanced learned predictive architecture. Multiple prediction learners responsible for different parts of the task and the controller combines them for use in signalling the user.}
    \label{fig:advanced_architecture}
\end{figure}

Three different look-ahead values were used in this study, which were manually chosen. One of those values worked with automated motion, but in the human trials further look-ahead, or perhaps adjusting the learning parameters, is required. It is not out of the question to think that different participants, or methods of delivery of the signal to the participant, would necessitate different amounts of look-ahead. In the vein of having the system learn about the world in terms referential to itself rather than having the designer impart specific task, or partner, knowledge upon it, a system that also learns the amount of look-ahead required may perform better and more generally. This could be done with multiple or layered predictions. An example of an architecture for such an interaction of prediction learners can be found in Figure n the supplementary material. In this proposed architecture, the robot interacting with the environment to generate predictions about contact remains the same. Following methods from Sherstan et al. \cite{Sherstan2015}, a second set of GFVs would learn about the way the arm is moving, and make predictions about what position the arm is expected to be in. These predictions can be combined in the controller to determine if the predicted position is in a predicted contact area, and this is in turn used to signal the user. This could enable a learned value in place of  what here is a designer-specified look ahead, and increase the success of the learning agent in assisting the human to avoid all contacts in future.

\end{document}